\pdfoutput=1

\documentclass[11pt]{article}

\usepackage[preprint]{acl}

\usepackage{times}
\usepackage{latexsym}

\usepackage[T1]{fontenc}

\usepackage[utf8]{inputenc}

\usepackage{microtype}

\usepackage{inconsolata}

\usepackage{graphicx}

\usepackage{microtype}
\usepackage{algorithm}
\usepackage{algpseudocode}
\usepackage{amsmath}
\usepackage{amsfonts}
\usepackage{amsthm}
\usepackage{amssymb}

\usepackage{setspace}
\usepackage{graphicx}
\usepackage{caption}
\usepackage{subcaption}
\usepackage{array,multirow}
\usepackage{bm}
\usepackage{adjustbox}
\usepackage{booktabs}
\usepackage{dirtytalk}

\title{A Collocation-based Method for Addressing Challenges \\ in Word-level Metric Differential Privacy}

\author{Stephen Meisenbacher, {\bf Maulik Chevli}, \and {\bf Florian Matthes} \\
  Technical University of Munich \\
  School of Computation, Information and Technology \\
  Department of Computer Science \\
  Garching, Germany \\
  \texttt{\{stephen.meisenbacher,maulikk.chevli,matthes\}@tum.de} \\
}

\begin{document}
\maketitle
\begin{abstract}
Applications of Differential Privacy (DP) in NLP must distinguish between the syntactic level on which a proposed mechanism operates, often taking the form of \textit{word-level} or \textit{document-level} privatization. Recently, several word-level \textit{Metric} Differential Privacy approaches have been proposed, which rely on this generalized DP notion for operating in word embedding spaces. These approaches, however, often fail to produce semantically coherent textual outputs, and their application at the sentence- or document-level is only possible by a basic composition of word perturbations. In this work, we strive to address these challenges by operating \textit{between} the word and sentence levels, namely with \textit{collocations}. By perturbing n-grams rather than single words, we devise a method where composed privatized outputs have higher semantic coherence and variable length. This is accomplished by constructing an embedding model based on frequently occurring word groups, in which unigram words co-exist with bi- and trigram collocations. We evaluate our method in utility and privacy tests, which make a clear case for tokenization strategies beyond the word level.
\end{abstract}

\section{Introduction}
The study of Differential Privacy (DP) in Natural Language Processing has brought about a number of innovative approaches, ranging from text rewriting to private fine-tuning of language models \cite{hu-etal-2024-differentially}. At the core of these approaches is the goal of providing a level of quantifiable privacy protection when text is shared or used for some downstream purpose. Among other advantages, leveraging DP allows for flexibility in choice of privacy level, governed by the privacy budget, or $\varepsilon$.

An early form of DP in NLP comes with the notion of \textit{word-level Metric Differential Privacy} (MLDP), the goal of which is to allow for privacy-preserving analysis on text documents by performing word-level \textit{perturbations} \cite{feyisetan_balle_2020}. In essence, a word is obfuscated by adding random noise to its embedding, perturbing to a (possibly different) word, and then releasing this \say{privatized} word \cite{klymenko-etal-2022-differential}. Metric DP is ensured via the implementation of \textit{mechanisms} which add calibrated noise to text representations. While other recent advances in DP NLP have shifted towards more complex language models, the simplicity and atomicity of word-level MLDP methods make a case for its further study.

Although these works show promising results in balancing privacy and utility in the MLDP setting, a number of challenges have also been highlighted \cite{klymenko-etal-2022-differential}. Firstly, the design of mechanisms raises challenges when the underlying spaces, e.g., word embeddings, are both vast (large vocabularies) and complex (high dimensional) \cite{feyisetan2021research}. Moreover, applying DP at the word level and composing these results for private text generation often results in texts with grammatical errors \cite{mattern-etal-2022-limits}. Beyond this, composed word-level MLDP will always lead to privatized documents with the same length as the input documents, diminishing privacy protections. 

In this work, we aim to address these challenges by building upon the promise of MLDP mechanisms, but rather than rely on \textit{word-level} perturbations, we extend these mechanisms to operate on the \textit{collocation-level}, or more generally, the \textit{n-gram} level. \textit{By specifically focusing on collocations, we hope to improve output text coherence, introduce generated length variability, and boost utility while also performing fewer overall perturbations, thus saving privacy budget.} In particular, we are guided by the following research question:

\begin{quote}
    \textit{Can collocations be leveraged to improve the function of word-level Metric Differential Privacy mechanisms, and what is the effect on privacy and utility?}
\end{quote}

We answer this question by designing a new approach for MDLP perturbations which leverages collocation embedding models in conjunction with two proposed collocation extraction algorithms. In our conducted utility and privacy tests, we show that this simple, yet meaningful augmentation leads to improved utility and comparable privacy under a number of privatization strategies. Concretely, the contributions of our work are as follows:
\begin{enumerate}
    \itemsep 0em
    \item To the best of the authors' knowledge, we are the first work to explore the use of collocations in the DP NLP space, most notably through the use of joint n-gram embedding models. 
    \item We demonstrate the effectiveness of using collocation-based embedding models as a basis for MLDP mechanisms, rather than previous word-level approaches.
    \item We provide a blueprint for further improving MLDP mechanisms through the open-sourcing of our collocation extraction algorithms and embedding models, found at \url{https://github.com/sjmeis/CLMLDP}.
\end{enumerate}

\section{Foundations}
\label{sec:foundations}
\subsection{Differential Privacy}
Differential Privacy (DP) \cite{dwork2006differential} provides mathematical privacy guarantees for individual's data when their data undergoes algorithmic processing. Intuitively, it provides plausible deniability on the result about the source of input to an algorithm. An algorithm (or a \textit{mechanism}) that is DP yields similar results irrespective of the inclusion of a single data record in the input dataset. These types of datasets that differ only in a single record are called \emph{adjacent} or \emph{neighboring} datasets.

Consider two adjacent datasets $D$ and $D'$ differing only in a single record. A randomized mechanism $\mathcal{M}: \mathcal{X}^m \to \mathcal{O}$ that takes a dataset $D \in \mathcal{X}^m$ and results in some output $O \in \mathcal{O} $ is called a $(\varepsilon, \delta)$-DP iff for all adjacent datasets $D, D'$ and $\forall O \subseteq \mathcal{O}$, the following holds with $\varepsilon \geq 0$ and $\delta \in [0, 1]$:

{\small
\begin{displaymath}
    \small \mathbb{P}[\mathcal{M}(\mathcal{D}) \in O] 
     \leq e^\varepsilon \cdot {\mathbb{P}[\mathcal{M}(\mathcal{D^\prime}) \in O]} + \delta
\end{displaymath}}

The notion of adjacency of datasets defines the element protected by DP. If adjacent datasets $D$ and $D'$ differ in one record, a DP mechanism provides plausible deniability about the inclusion or exclusion of a single record in the dataset. When the data records are collected at a central location and then a DP mechanism is to be applied, the adjacency notion can be defined as aforementioned and it is called \textit{Global DP}. However, if the data collector is not trusted and the DP mechanism is applied locally before the collection of data, the notion of adjacency is defined as any two data records; this is called \textit{Local DP} \cite{6686179}. 

For natural language, the unstructured nature of data brings additional challenges regarding the notion of adjacent datasets \cite{klymenko-etal-2022-differential}. We consider a text consisting of $n$-gram tokens, and define the notion of adjacency as any two tokens following \citet{feyisetan_balle_2020}. Hence, an adversary cannot determine with high probability the source token of the privatized token. 

\subsection{Metric Differential Privacy (MDP)}
For two finite sets $\mathcal{X}$ and $\mathcal{Z}$ and a distance metric $d: \mathcal{X} \times \mathcal{X} \to \mathbb{R}+$ defined for the set $\mathcal{X}$, a randomized mechanism $\mathcal{M}: \mathcal{X} \to \mathcal{Z}$ satisfies metric differential privacy or $\varepsilon d_{\mathcal{X}}$-privacy \textit{iff} $\forall x, x' \in \mathcal{X}$ and $\forall z \in \mathcal{Z}$, this condition is satisfied with $\varepsilon > 0$:

{\footnotesize
	\begin{equation}
            \small
		\frac{ \mathbb{P}[ \mathcal{M}(x) = z ] }
			 { \mathbb{P}[ \mathcal{M}(x^\prime) = z ] } 
		\leq e^{\varepsilon d(x, x^\prime)}
		\label{eq: mdp}
	\end{equation}}

Metric DP is a relaxation of DP where instead of considering the worst-case guarantees, the privacy guarantees scale according to the distance between adjacent datasets \cite{10.1007/978-3-642-39077-7_5}. This allows for greater utility and flexibility alongside a mathematical guarantee.

\subsection{MDP for a Sentence}
\label{sec:sentence}
We assume a vocabulary set consisting of all the tokens in $\mathcal{V}$, with the tokens as points in the embedding space. The embedding function $\Phi: \mathcal{V} \to \mathbb{R}^d$ gives the position of the tokens in the space. Additionally, we assume that the space $\mathcal{V}$ is equipped with a distance metric $d_\mathcal{V}: \mathcal{V} \times \mathcal{V} \to \mathbb{R}+$ that gives us the distance between two tokens $w$ and $w^\prime$ as 
\begin{equation}
    \small d_\mathcal{V}(w, w^\prime) = || \Phi(w) - \Phi(w') ||_2
    \label{eq: distance}
\end{equation}

If a mechanism $\mathcal{M}$ satisfies MDP for two tokens for $\varepsilon > 0$, it satisfies Equation \ref{eq: mdp} $\forall w, w^\prime \in \mathcal{V}$, and thus, we have the following inequality:
\begin{equation}
    \small
    \frac{ \mathbb{P}[ \mathcal{M}(w) = x ] }
			 { \mathbb{P}[ \mathcal{M}(w^\prime) = x ] }
    \leq e ^ {\varepsilon \cdot d_V(w, w^\prime)}
\end{equation}

This guarantee can be extended to the whole sentence consisting of $n$ tokens, i.e., $s = w_1\cdot w_2\cdots w_n$. Following \citet{feyisetan_balle_2020}, a token-level mechanism can be applied to each token independently and a privatized sentence can be generated by concatenating these privatized tokens, i.e., $z = x_1\cdot x_2 \cdots x_n$. If the distance function that takes sentences of the same token length $D: \mathcal{V}^n \times \mathcal{V}^n \to \mathbb{R+}$ is defined as $\tiny\mathcal{D} = \sum_{i=1}^n d_\mathcal{V}(w_i, x_i)$, the privacy guarantees of applying mechanism $\mathcal{M}$ to the sentence can be derived as follows:

{\small \vspace{-1em} \begin{align*}
    \frac{\mathbb{P}[\mathcal{M}(s) = z]}
             {\mathbb{P}[\mathcal{M}(s^\prime) = z]}
        &=  \displaystyle\prod_{i=1}^n \frac{\mathbb{P}[\mathcal{M}(w_i) = x]}{\mathbb{P}[\mathcal{M}(w^\prime_i) = x]}
         \\
        &\leq \displaystyle\prod_{i=1}^n \exp{ \left( \varepsilon \cdot d_\mathcal{V}(w_i, w_i^\prime) \right) }
        \\
        &=\exp{ \left( \varepsilon \cdot  \displaystyle\sum_{i=1}^nd_\mathcal{V}(w_i, w_i^\prime) \right) }
        \\
        &=\exp{ \left( \varepsilon \cdot \mathcal{D}(s, s^\prime) \right)}
\end{align*} }%

It should be noted that while we use the term \say{sentence} here, the above can be generalized to text \say{documents}.

\subsection{The Theory of Collocations}
In linguistics, \textit{collocations} are defined as groupings of words that often appear together in language. More specifically, collocations are word groups (\say{multi-word expressions}) existing in the space between idioms and free word groups \cite{mckeown2000collocations}, where the meaning of idioms cannot be understood by their individual words.
Intuitively, collocations can be defined as groupings of words that appear in predictable patterns (\textit{good morning}), without being as rigid as idioms (\textit{sleep like a baby}) \cite{mckeown2000collocations}.


An important concept is the \textit{Contextual Theory of Meaning} of John Rupert Firth \cite{leon2005meaning,manning1999foundations}, famously summarized by \say{a word is characterized by the company it keeps}. The meaning of a given collocation only takes form when viewing the group as a whole, and not by examining the meaning of each word individually. 

Looking to the notion of differentially private text rewriting via the composition of word-level replacements, one may imagine that the theory of collocations sheds light on the potential pitfalls of isolated word substitutions. As highlighted by \citet{mattern-etal-2022-limits}, word-level DP disregards context, which results in semantically disjoint replacements as well as frequent grammatical incongruities. In this light, we posit that collocations may improve both of these challenges, as collocations represent groups of words with \textit{bundled} meaning, and within a collocation, proper grammar must be upheld.

\section{Related Work}

\subsection{Word-level MLDP}
While \citet{fernandes2019generalised} proposed an early implementation of metric DP, \cite{feyisetan_balle_2020} were the first to design a word-level MLDP mechanism for static word embeddings. Ensuing works aim to improve word-level methods through various means, including differing metrics \cite{xu2020differentially}, nearest neighbor mapping \cite{xu2021utilitarian,meisenbacher20241diffractor}, or noise mechanism \cite{xu2021density,carvalho2023tem}. Other works focus on the selection of words to privatize \cite{yue,Chen2022ACT}.

We aim to build upon this body of work, while also addressing the known challenges of semantic coherence, grammatical correctness, and output text length variability. In particular, we tackle these challenges in the word-level MLDP setting by leveraging \textit{collocations} and \textit{n-gram embeddings}.

\begin{figure*}[htbp]
    \centering
    \includegraphics[scale=0.55]{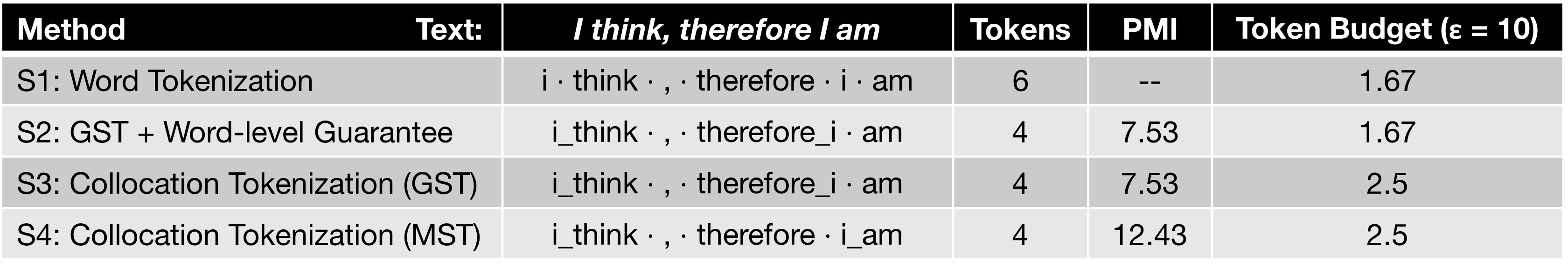} 
    \vspace{-0.4em}
    \caption{An example of word tokenization versus collocation tokenization. Collocation tokenization will often result in fewer tokens, as collocations frequently occur in natural language. \textit{Token budget} denotes the privacy budget assigned to each token given an example document-level budget (e.g., $\varepsilon = 10$) and assuming basic composition.}
    \vspace{-0.9em}
    \label{fig:example}
\end{figure*}

\subsection{Collocation Extraction and Evaluation}
Several computational approaches for automatic collocation extraction have been explored. \citet{pecina-2005-extensive} surveys an extensive list of early collocation extraction methods, and later explores the combination of different metrics \cite{pecina-schlesinger-2006-combining}. Other works improve on classic association measures \cite{bouma-2010-collocation,jbp:/content/journals/10.1075/ijcl.20.2.01bre}, and more recent work has focused on evaluating end-to-end solutions \cite{bhalla-klimcikova-2019-evaluation, espinosa-anke-etal-2021-evaluating}. More on the theoretical underpinnings and our motivation for the use of collocations can be found in Section \ref{sec:collocations}.

\subsection{N-gram Embeddings}
Extending static embedding models beyond the word level often takes the form of \textit{n-gram} embeddings or \textit{phrase} embeddings \cite{poliak-etal-2017-efficient,yin-schutze-2014-exploration}.  Works have explored different methods of embedding n-grams, notably the use of Pointwise Mutual Information (PMI) \cite{zhao-etal-2017-ngram2vec} or BERT-based models for more contextual phrase embeddings \cite{wang-etal-2021-phrase}.

In a study of n-gram embeddings, \citet{gupta-etal-2019-better} find that the joint training process improves the quality of single-word embeddings. In other works, it is shown that n-gram embeddings can improve a variety of NLP tasks \citet{10.1145/3219819.3219897,zhang-etal-2014-bilingually, yin-schutze-2015-discriminative}.

With these works as motivation, we investigate whether n-gram embeddings can serve to improve DP text privatization approaches previously relying on word embeddings. In particular, we explore the usefulness of embedding \textit{collocations} as the underlying embedding model of MLDP mechanisms.

\section{A Collocation-based MLDP Method}
\label{sec:collocations}
In this section, we describe our proposed method, which differs from word-level MLDP methods in that it sets the underlying metric space to that of a \textit{jointly trained} model of unigrams, bigram collocations, and trigram collocations. We outline a method to extract collocations, the training of the abovementioned embedding model, and the augmentation of existing MLDP mechanisms.

\subsection{Extracting Collocations}
The first challenge of dealing with collocations is the reliable extraction of meaningful multi-word expressions that uphold the definition of a collocation. Several methods have been proposed by the literature, ranging from simple frequency-based approaches, methods looking at syntactic co-occurrences, to \textit{hypothesis testing} methods or \textit{association measures} such as mutual information \cite{Evert+2009+1212+1248, manning1999foundations}.

In this work, we focus on the extraction of bigram and trigram collocations via the use of \textit{Pointwise Mutual Information} (PMI) \cite{church-hanks-1990-word}. Essentially, PMI indicates how much one point (word) tells us about another. In other words, if the presence of one word \textit{decreases} the uncertainty of the presence of another word, these two words have a high PMI. In the case of bigrams, two words $x$ and $y$ have a PMI as follows:
\begin{equation}
    \label{eq:pmi}
    PMI(x,y) = \log_2 \frac{P(x\vert y)}{P(x)} = \log_2 \frac{P(y\vert x)}{P(y)}
\end{equation}

Given a corpus of $N$ words, we can empirically measure the bigram PMI of $xy$ as defined in Equation \ref{eq:pmi} by the following:
\begin{equation}
    \label{eq:bipmi}
    PMI(x,y) = \log_2 \frac{N \cdot c(xy)}{c(x) \cdot c(y)}
\end{equation}

Note that in Equation \ref{eq:bipmi}, the order of the unigrams matters, and $c$ denotes the raw frequency count of a given unigram or bigram. For trigram collocations, a simple modification can be made:
\begin{equation}
    \label{eq:tripmi}
    PMI(x,y,z) = \log_2 \frac{N^2 \cdot c(xyz)}{c(x) \cdot c(y) \cdot c(z)}
\end{equation}

\subsubsection{Empirical Collocations}
For the extraction of \textit{empirical} collocations \cite{Evert+2009+1212+1248}, i.e., those that can be derived via empirical means, we measure the PMI of bigrams and trigrams from a selected random sample of 2.5 million texts of the publicly available large-scale text corpus \textsc{C4} (Colossal Cleaned Common Crawl) \cite{JMLR:v21:20-074}. After counting the frequency of all unigrams, bigrams, and trigrams, we calculate the bigram and trigram PMI values using Equations \ref{eq:bipmi} and \ref{eq:tripmi}, respectively. We filter the results for all values with a PMI score of 2.0 or higher \textit{and} not containing any English connector words (e.g., \textit{a}, \textit{an}, \textit{the}, \textit{and}, \textit{or}, etc.)\footnote{As defined by the Python \textsc{gensim} package.}. This process results in a set of 3.02 million bigrams and 1.31 million trigrams\footnote{Can be found in the data folder of our code repository.}.

\subsubsection{Collocation-level Tokenization}
We design an extraction algorithm that will tokenize a given input text into its unigram, bigram, and trigram counterparts based upon the empirically derived PMI scores of the collocations. To do this, we define two scoring methods (pseudocode found in Appendix Algorithms \ref{alg:gst} and \ref{alg:mst}):

\begin{itemize}
    \itemsep -0.1em
    \item \textbf{Greedy Sequential Tokenization (GST)}: a text is tokenized \textit{greedily} by processing the tokens in order, with trigrams being prioritized. This is described in Algorithm \ref{alg:gst}.
    \item \textbf{Max Score Tokenization (MST)}: a text is tokenized in a way that maximizes the overall PMI score of the resulting tokenized text. This is described in Algorithm \ref{alg:mst}. 
\end{itemize}

\begin{algorithm}[htbp]
\caption{\newline \small Greedy Sequential Tokenization (GST)}
\label{alg:gst}
    \begin{algorithmic}
        \footnotesize
        \Require scored bigrams $B$, scored trigrams $T$, input $text$

        \State $\texttt{tkns} \gets word\_tokenize(text)$
        \State $\texttt{bigram\_cands} \gets \textit{get\_bigrams}(\texttt{tkns}).\textit{intersect}(B)$
        \State $\texttt{trigram\_cands} \gets \textit{get\_trigrams}(\texttt{tkns}).\textit{intersect}(T)$
        \State $\texttt{n} \gets \textit{length}(\texttt{tkns})$

        \State $\texttt{output} \gets []$
        \For {$idx \in 1...n$}
            \State $\texttt{cand} \gets \texttt{trigram\_cands}.\textit{find}(\texttt{tkns}[idx\colon idx+2])$
            \If {$!\texttt{cand}$}
                \State $\texttt{cand} \gets \texttt{bigram\_cands}.\textit{find}(\texttt{tkns}[idx\colon idx+1])$
            \EndIf
            \If {$!\texttt{cand}$}
                \State $\texttt{output}.append(text[idx]) \ \ \ \ $ \Comment{unigram}
            \Else
                \State $\texttt{output}.append(\texttt{cand})$
            \EndIf
            \State $\texttt{bigram\_cands}.delete(\texttt{cand})$
            \State $\texttt{trigram\_cands}.delete(\texttt{cand})$
            \If {$\texttt{cand} \in B$} \Comment{advance to next unmatched word}
                \State $\texttt{idx} \mathrel{+}= 2$ 
            \Else
                \State $\texttt{idx} \mathrel{+}= 3$ 
            \EndIf
            
        \EndFor
        \State \Return $\texttt{output}$
    \end{algorithmic}
\end{algorithm}

\begin{algorithm}[ht!]
\caption{\newline \small Max Score Tokenization (MST)}
\label{alg:mst}
    \begin{algorithmic}
        \footnotesize
        \Require scored bigrams $B$, scored trigrams $T$, input $text$

        \State $\texttt{unigrams} \gets word\_tokenize(text)$
        \State $\texttt{bigram\_cands} \gets \textit{get\_bigrams}(text).\textit{intersect}(B)$
        \State $\texttt{trigram\_cands} \gets \textit{get\_trigrams}(text).\textit{intersect}(T)$
        
        \State $\texttt{cands} \gets sorted(\texttt{unigrams}\mathrel{+}\texttt{bigram\_cands}\mathrel{+}\texttt{trigram\_cands})$ 
        \State $\texttt{n} \gets \textit{length}(\texttt{cands})$
        \State $\texttt{matched} \gets []$
        \State $\texttt{output} \gets []$
        \For {$idx \in 1...n$}
            \If{$all(\texttt{cands}.tokens \mathrel{!}\in \texttt{matched})$}
                \State $\texttt{output}.append(\texttt{cand}[idx])$
                \State $\texttt{matched}.add(\texttt{cands}.tokens)$
            \EndIf
        \EndFor
        \State \Return $\texttt{output}$
    \end{algorithmic}
\end{algorithm}

GST and MST output a list of \say{tokens}, which can be either unigrams, bigram collocations, or trigram collocations. In its application, we tokenize documents at the \textit{sentence-level}, so as not to detect collocations across sentence boundaries. Note that this method can be extended to an arbitrary $n$-gram level. As a result, there are collocation tokens less than or equal to the number of word tokens.

\subsection{A Collocation Embedding Space}
We train an embedding model in which unigram words, bigram collocations, and trigram collocations co-exist in a single embedding space. In particular, we train a 300-dimension \textsc{Word2vec} model \cite{mikolov2013efficient} using the \textsc{gensim} package \cite{rehurek_lrec}.

To train the model, we leverage a large subset of the C4 Corpus, namely 250 million text samples, or roughly 500GB. As inputs to the \textsc{gensim} trainer, we give the text samples as tokenized by our two algorithms, namely GST and MST, thus resulting in two trained embedding models. The models were trained on a six-core Intel Xeon CPU, with the entire training process (extraction + embedding) taking roughly 90 hours per model. These models are made available in our code repository.

\subsection{Augmenting MLDP Mechanisms}
With the two collocation embedding models, we can now make a simple augmentation to existing word-level MLDP mechanisms. As these mechanisms typically operate on strictly word (unigram) spaces, we first swap out these models with our trained embedding models. Then, inputs to the mechanisms are tokenized by our collocation extraction algorithms, rather than word tokenization. 

The returned tokens can be of word length 1-3. However, the MLDP privacy guarantees are not affected, as the embedding space consists of these variable word-length tokens. Hence, the mechanisms can operate as usual, with the outputs being perturbed uni-, bi-, \textit{or} trigrams. Mathematically, the privacy guarantees for any tokens $w, w^\prime$ in our embedding space remain as defined in Section \ref{sec:sentence}.

\section{Experimental Setup and Results}
In our experiments to test our collocation-based method, we focus on evaluating the effect that can be observed by using collocations rather than pure words. In particular, we perform a two-part evaluation: utility experiments and privacy experiments.

\subsection{Mechanism Selection}
We center our evaluation around the fundamental MLDP mechanism proposed by \citet{feyisetan_balle_2020}, often referred to as \textsc{MADLIB} (Algorithm \ref{alg:madlib}), which typically operates on word embeddings in Euclidean space by adding calibrated multivariate noise. Our goal is to experiment using this mechanism across a range of $\varepsilon$ values, with the hopes of generalizing to mechanisms that build on top of \textsc{MADLIB}. Specifically, we choose the values $\varepsilon \in \{0.1,0.5,1,5,10,15,25,50\}$.

\begin{algorithm}[ht!]
\small
\caption{\newline \small MADLIB \cite{feyisetan_balle_2020}}
\begin{algorithmic}
\Require String $x = w_1 w_2 \ldots w_n$, privacy parameter $\epsilon > 0$, word set $\mathcal{W}$, embedding function $\varphi$
\Ensure Privatized string $\hat{x}$
\For{$i \in \{1, \ldots, n\}$}
\State Compute embedding $\varphi_i = \varphi(w_i)$ 
\State \parbox[t]{195pt}{Perturb embedding to obtain $\hat{\varphi}_i = \varphi_i + \mathcal{N}$ with noise density $p_\mathcal{N}(z) \propto \exp(-\epsilon \Vert z\Vert)$ \strut}
\State Obtain perturbed word $\hat{w}_i = \arg\min_{u\in \mathcal{W}} |\varphi(u) - \hat{\varphi}_i|$ 
\State Insert $\hat{w}_i$ in $i^{th}$ position of $\hat{x}$
\EndFor
\State \textbf{return} $\hat{x}$
\end{algorithmic}
\label{alg:madlib}
\end{algorithm}

\subsection{Utility Experiments}
\label{sec:util}
Our utility experiments follow the example set by several previous DP NLP works \cite{mattern-etal-2022-limits,utpala-etal-2023-locally,igamberdiev-habernal-2023-dp}, that is to evaluate how well DP generated text can preserve the original utility of the dataset. In particular, texts that are generated by a mechanism are compared against a non-privatized baseline, and the utility (loss) is measured.

To ensure a greater practical relevance, we perform utility experiments for our chosen mechanism at a \textit{document level}, where privatized documents are achieved via the composition of token-level perturbations. For this, we set a \textit{dataset specific privacy budget}, where our \say{base} $\varepsilon$ values introduced above are scaled by the average word length of each dataset. Thus, each text is perturbed with an overall budget of $\varepsilon * \textit{avg\_word\_len}(dataset)$. This ensures that all texts, regardless of length, are offered the same privacy guarantee. 

We note here that in this budget calculation, our goal is to provide an equal guarantee for each document to be privatized. However, we do not take into account the effect of the distance function in the Metric DP guarantee; thus, the document level budget is calculated according to pure DP composition, namely with basic composition of $\varepsilon$ values.

We evaluate five privatization strategies, which are described below and illustrated in Figure \ref{fig:example}:
\begin{enumerate}
    \itemsep 0em
    \item \textbf{Non-private}: no DP is applied to a given text.
    \item \textbf{Word-level} (S1): a text is tokenized by \textit{word}, and the document budget is distributed evenly to each word to be perturbed. For embeddings, we use \textsc{word2vec-google-news-300}\footnote{\url{https://code.google.com/archive/p/word2vec/}}. Since this model contains three billion tokens, we filter the vocabulary down to that of the \textsc{deberta-v3-base} (see next section). In S1, stopwords are not privatized.
    \item \textbf{Collocation-level, word-level guarantee} (S2): a text is tokenized using our GST collocation extraction algorithm, but each resulting \textit{token} is given the same budget as in the \textbf{word-level} scenario (see Figure \ref{fig:example}).
    \item \textbf{Collocation-level (GST)} (S3): a text is tokenized by GST, and the document budget is distributed evenly to all resulting collocations.
    \item \textbf{Collocation-level (MST)} (S4): same as above, but with the MST algorithm.
\end{enumerate}

Thus, for each given input text, we receive five resulting outputs: the original (baseline) text and four privatized variants. These serve as the basis for our utility (and privacy) experiments.

\subsection{Training and Evaluation}
\paragraph{Datasets}
To measure utility, we choose four datasets from the \textsc{GLUE} benchmark \cite{wang-etal-2018-glue}, a standard benchmark representing a variety of language understanding tasks. Specifically, we utilize the \textsc{CoLA}, \textsc{MRPC}, \textsc{RTE}, and \textsc{SST2} datasets. For\textsc{SST2}, we use a 10k random sample.

We first perturb each dataset according to the strategies outlined above. Note that we privatize both the train and validation sets, as this presents the strictest test of utility preservation in which all data is perturbed. For datasets with two sentences (\textsc{RTE}, \textsc{MRPC}), we only perturb the first sentence.

\paragraph{Model Training}
The preservation of utility is measured by fine-tuning a language model on all dataset variants (i.e., baseline or perturbed), and measuring the effect on utility. For this, we fine-tune all datasets on a \textsc{deberta-v3-base} model with input size of $256$, for one epoch and otherwise default HuggingFace Trainer parameters. All training is performed on one NVIDIA A6000 GPU. For stability in the results, we run each training procedure three times on different random shuffles of the data, reporting the average metrics of all runs.

\paragraph{Metrics}
We report the (micro) F1 score of all trained models on the validation sets. This metric aims to capture the effect of privatization on the ability for a model with good utility to be trained.

In addition, we report the \textit{cosine similarity} (CS) between each (\textit{original}, \textit{private}) dataset pair. This metric can be used to measure the degree to which semantic similarity is preserved in perturbation \cite{meisenbacher2024comparative}. For this, we utilize the \textsc{sentence-transformers/all-MiniLM-L6-v2} model \cite{reimers-2019-sentence-bert}.

We also use \textit{perplexity} to measure the semantic coherence privatized texts. As perplexity aims to measure the ability of a language model to predict a given text, a better (lower) perplexity would imply a text is more \say{natural} or \say{predictable}. Although this metric has been used in recent DP NLP works \cite{yue-etal-2023-synthetic,singh2024whispered}, its use directly on privatized texts has not been explored widely with the exception of \citet{10.1145/3485447.3512232}. We report \textit{average perplexity} (AP) of all sequences in a dataset, using \textsc{GPT-2} \cite{radford2019language}.

\subsection{Privacy Experiments}
Our privacy experiments take the form of \textit{empirical privacy} measurement, where we use two tasks as proxies for privacy preservation, which also allow for measures of \textit{relative gain} (discussed below):

\begin{enumerate}
    \itemsep 0em
    \item \textbf{Yelp Reviews} \cite{NIPS2015_250cf8b5}: we utilize the same dataset used by \citet{utpala-etal-2023-locally}, which contains a subset of reviews authored by 10 frequent reviewers. From this, we model an \textit{authorship identification} task. We take a random subset of 10k rows.
    \item \textbf{Trustpilot Reviews} \cite{10.1145/2736277.2741141}: each review includes the gender of the original reviewer (M/F). This creates an \textit{gender identification} task, for which we use a 10k sample.
\end{enumerate}

As with the utility experiments, all texts in the two datasets are privatized according to the five perturbation strategies. The resulting datasets are then divided into a 90-10 train-validation split\footnote{A random seed of \textit{42} is used throughout this work.}.

\paragraph{Evaluation}
Both datasets are labeled for sentiment (positive/negative), allowing for a binary classification task, which is carried out in a similar manner as the utility experiments. Macro F1 is reported, as the labels are positive-biased. 

Next, empirical privacy is measured. To do this, an adversarial classifier is trained to predict the sensitive attribute (author ID or gender) given the corresponding text. We use the same \textsc{deberta-v3-base} fine-tuning process for the creation of this classifier. For evaluation, we follow two adversarial archetypes as proposed in the recent literature \cite{mattern-etal-2022-limits, utpala-etal-2023-locally}: the \textit{static} and \textit{adaptive} attackers. The static attacker is only able to train on the non-privatized train split and must evaluate on privatized validation splits. The adaptive attacker, a much more capable adversary, is able to train on the privatized train splits.

For adversarial performance, we report macro F1 scores. Using both the utility and privacy measurements, we calculate the \textit{relative gain} (RG) of privatization \cite{mattern-etal-2022-limits}, namely whether the gains in privacy outweigh potential losses in utility. This metric is given by the following formula, where $P_p$, $U_p$, $P_o$, $U_o$ are the measured privacy (\textit{P}) and utility (\textit{U}) scores of the privatized ($_p$) or original ($_o$) data: $RG = (U_p/ U_o) - (P_p / P_o)$.

\subsection{Results}
\begin{figure}[ht!]
    \centering
    \includegraphics[scale=0.5]{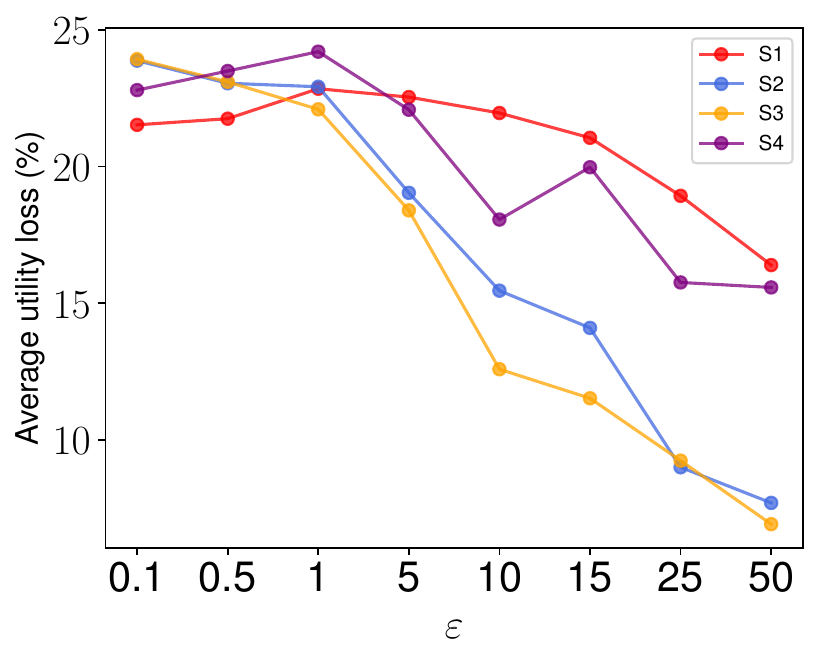}
    \vspace{-10pt}
    \caption{Average Utility Loss. This graph depicts the average utility loss (in F1) for a given base $\varepsilon$ value across four \textsc{GLUE} tasks and our four privatization strategies.}
    \label{fig:drop}
\end{figure}

The results of the utility experiments are given in Tables \ref{tab:cs_1column}, \ref{tab:ap_1column}, and \ref{tab:f1}, and are illustrated in Figure \ref{fig:drop}, whereas the privacy results are shown in Table \ref{tab:ep}.

\begin{table}[ht!]
    \centering
    \resizebox{\linewidth}{!}{
\begin{tabular}[t]{c|llllllll}
\multicolumn{1}{r|}{$\varepsilon$} & \multicolumn{1}{c}{0.1} & \multicolumn{1}{c}{0.5} & \multicolumn{1}{c}{1} & \multicolumn{1}{c}{5} & \multicolumn{1}{c}{10} & \multicolumn{1}{c}{15} & \multicolumn{1}{c}{25} & \multicolumn{1}{c}{50} \\ \hline
S1 & 0.13 & 0.13 & 0.13 & 0.14 & 0.18 & 0.25 & 0.38 & 0.63 \\
S2 & 0.16 & 0.16 & 0.18 & 0.42 & 0.65 & 0.78 & 0.88 & 0.94 \\
S3 & 0.16 & \textbf{0.17} & \textbf{0.20} & \textbf{0.51} & \textbf{0.74} & \textbf{0.85} & \textbf{0.92} & \textbf{0.96} \\
S4 & \textbf{0.17} & 0.15 & 0.19 & 0.33 & 0.45 & 0.52 & 0.60 & 0.68
\end{tabular}
}
\caption{Average cosine similarity between original and privatized texts across all four utility datasets.}
\label{tab:cs_1column}
\vspace{-10pt}
\end{table}

\begin{table}[ht!]
    \centering
    \resizebox{\linewidth}{!}{
\begin{tabular}[t]{c|llllllll}
Baseline & \multicolumn{8}{c}{622} \\ \hline
\multicolumn{1}{r|}{$\varepsilon$} & \multicolumn{1}{c}{0.1} & \multicolumn{1}{c}{0.5} & \multicolumn{1}{c}{1} & \multicolumn{1}{c}{5} & \multicolumn{1}{c}{10} & \multicolumn{1}{c}{15} & \multicolumn{1}{c}{25} & \multicolumn{1}{c}{50} \\ \hline
S1 & \textbf{1731} & \textbf{1967} & \textbf{2325} & \textbf{3593} & 5150 & 5525 & 5978 & 3987 \\
S2 & 3913 & 4135 & 4774 & 4037 & 2953 & 2239 & 1714 & 1582 \\
S3 & 3848 & 4237 & 4960 & 3609 & \textbf{2418} & \textbf{1925} & \textbf{1632} & \textbf{1547} \\
S4 & 4855 & 5456 & 6103 & 5429 & 4673 & 3056 & 2574 & 2302
\end{tabular}
}
\caption{Average perplexity of the privatized texts across all four utility datasets, where lower scores are better.}
\label{tab:ap_1column}
\vspace{-10pt}
\end{table}

\begin{table*}[htbp]
    \begin{subtable}[t]{0.5\textwidth}
    \centering
    \resizebox{0.99\linewidth}{!}{%
        \begin{tabular}{c|llllllll}
Baseline & \multicolumn{8}{c}{$84.97_{0.4}$} \\ \hline
$\varepsilon$ & \multicolumn{1}{c}{0.1} & \multicolumn{1}{c}{0.5} & \multicolumn{1}{c}{1} & \multicolumn{1}{c}{5} & \multicolumn{1}{c}{10} & \multicolumn{1}{c}{15} & \multicolumn{1}{c}{25} & \multicolumn{1}{c}{50} \\ \hline
S1 &  $69.13_{0.0}$ & $69.13_{0.0}$ & $69.13_{0.0}$ & $69.13_{0.0}$ & $69.13_{0.0}$ & $69.13_{0.0}$ & $69.13_{0.0}$ & $69.13_{0.0}$  \\
S2 & $69.13_{0.0}$ & $69.13_{0.0}$ & $69.13_{0.0}$ & $69.13_{0.0}$ & $72.83_{1.3}$ & $74.11_{1.0}$ & $78.17_{0.0}$ & $79.42_{0.2}$ \\
S3 & $69.13_{0.0}$ & $69.13_{0.0}$ & $69.13_{0.0}$ & $69.13_{0.0}$ & $\mathbf{73.27_{1.9}}$ & $\mathbf{75.01_{1.1}}$ & $\mathbf{80.22_{0.9}}$ & $\mathbf{81.85_{0.4}}$ \\
S4 & $69.13_{0.0}$ & $69.13_{0.0}$ & $69.13_{0.0}$ & $69.13_{0.0}$ & $69.13_{0.0}$ & $69.13_{0.0}$ & $69.16_{0.0}$ & $69.13_{0.0}$
\end{tabular}
    }
    \caption{CoLA (Avg. Words/Text: 7.80)}
    \label{tab:f1_cola}
    \end{subtable}
    \hfill
    \begin{subtable}[t]{0.5\textwidth}
    \centering
    \resizebox{0.99\linewidth}{!}{%
        \begin{tabular}{c|llllllll}
Baseline & \multicolumn{8}{c}{$94.33_{0.2}$} \\ \hline
$\varepsilon$ & \multicolumn{1}{c}{0.1} & \multicolumn{1}{c}{0.5} & \multicolumn{1}{c}{1} & \multicolumn{1}{c}{5} & \multicolumn{1}{c}{10} & \multicolumn{1}{c}{15} & \multicolumn{1}{c}{25} & \multicolumn{1}{c}{50} \\ \hline
S1 & $\mathbf{58.75_{1.9}}$ & $\mathbf{56.0_{3.6}}$ & $53.94_{2.9}$ & $56.8_{0.6}$ & $56.73_{5.1}$ & $58.87_{2.4}$ & $67.78_{1.8}$ & $76.11_{0.8}$  \\
S2 & $50.76_{0.2}$ & $50.92_{0.0}$ & $53.25_{1.7}$ & $68.0_{0.4}$ & $79.05_{0.9}$ & $82.76_{0.4}$ & $91.67_{0.5}$ & $\mathbf{93.16_{0.7}}$ \\
S3 & $50.92_{0.0}$ & $52.22_{1.8}$ & $\mathbf{56.15_{0.7}}$ & $\mathbf{71.18_{0.6}}$ & $\mathbf{84.56_{1.0}}$ & $\mathbf{87.69_{0.4}}$ & $\mathbf{92.51_{0.4}}$ & $92.78_{0.4}$ \\
S4 & $51.61_{0.3}$ & $50.92_{0.0}$ & $52.68_{2.5}$ & $57.11_{4.8}$ & $71.25_{0.5}$ & $65.9_{0.8}$ & $80.2_{2.1}$ & $80.24_{0.4}$
\end{tabular}
    }
        \caption{SST2 (Avg. Words/Text: 8.82)}
        \label{tab:f1_sst2}
    \end{subtable}
    \hfill
    \begin{subtable}[t]{0.5\textwidth}
    \centering
    \resizebox{\linewidth}{!}{%
        \begin{tabular}{c|llllllll}
Baseline & \multicolumn{8}{c}{$85.34_{1.0}$} \\ \hline
$\varepsilon$ & \multicolumn{1}{c}{0.1} & \multicolumn{1}{c}{0.5} & \multicolumn{1}{c}{1} & \multicolumn{1}{c}{5} & \multicolumn{1}{c}{10} & \multicolumn{1}{c}{15} & \multicolumn{1}{c}{25} & \multicolumn{1}{c}{50} \\ \hline
S1 & $69.28_{0.8}$ & $69.93_{1.2}$ & $\mathbf{70.02_{0.5}}$ & $68.38_{0.0}$ & $69.69_{0.6}$ & $70.1_{0.2}$ & $70.75_{0.1}$ & $70.75_{0.5}$ \\
S2 & $69.93_{1.2}$ & $\mathbf{70.67_{0.4}}$ & $69.2_{1.2}$ & $\mathbf{69.85_{1.1}}$ & $70.26_{1.3}$ & $71.08_{2.3}$ & $\mathbf{76.8_{4.5}}$ & $80.72_{1.7}$ \\
S3 & $69.2_{1.2}$ & $69.53_{1.6}$ & $69.61_{1.0}$ & $69.12_{1.0}$ & $\mathbf{74.35_{1.6}}$ & $\mathbf{73.37_{2.5}}$ & $74.75_{4.6}$ & $\mathbf{81.29_{1.0}}$ \\
S4 & $\mathbf{70.26_{1.3}}$ & $69.12_{1.0}$ & $68.38_{0.0}$ & $69.44_{1.2}$ & $71.24_{0.1}$ & $70.02_{1.2}$ & $72.06_{1.1}$ & $71.81_{2.1}$
\end{tabular}
    }
        \caption{MRPC (Avg. Words/Text: 19.54)}
        \label{tab:f1_mrpc}
    \end{subtable}
    \hfill
    \begin{subtable}[t]{0.5\textwidth}
    \centering
    \resizebox{0.99\linewidth}{!}{%
        \begin{tabular}{c|llllllll}
Baseline & \multicolumn{8}{c}{$70.97_{2.0}$} \\ \hline
$\varepsilon$ & \multicolumn{1}{c}{0.1} & \multicolumn{1}{c}{0.5} & \multicolumn{1}{c}{1} & \multicolumn{1}{c}{5} & \multicolumn{1}{c}{10} & \multicolumn{1}{c}{15} & \multicolumn{1}{c}{25} & \multicolumn{1}{c}{50} \\ \hline
S1 & $52.35_{0.5}$ & $\mathbf{53.55_{0.7}}$ & $51.14_{3.0}$ & $51.14_{2.2}$ & $52.23_{0.5}$ & $53.31_{1.4}$ & $52.23_{0.9}$ & $\mathbf{54.03_{1.5}}$ \\
S2 & $50.3_{3.4}$ & $52.71_{0.0}$ & $52.35_{0.5}$ & $52.47_{0.6}$ & $51.62_{3.5}$ & $51.26_{1.2}$ & $\mathbf{52.95_{2.1}}$ & $51.5_{1.2}$ \\
S3 & $50.66_{2.9}$ & $52.35_{0.8}$ & $\mathbf{52.35_{0.3}}$ & $\mathbf{52.59_{0.2}}$ & $\mathbf{53.07_{3.1}}$ & $\mathbf{53.43_{2.3}}$ & $51.14_{1.3}$ & $51.99_{0.8}$ \\
S4 & $\mathbf{53.43_{1.0}}$ & $52.47_{0.3}$ & $48.62_{3.0}$ & $51.62_{1.5}$ & $51.74_{1.9}$ & $50.66_{2.2}$ & $51.14_{3.0}$ & $52.11_{3.4}$
\end{tabular}
    }
        \caption{RTE (Avg. Words/Text: 44.48)}
        \label{tab:f1_rte}
    \end{subtable}
    \caption{Utility Experiment Results. All results represent average micro F1 scores over three training runs, with the standard deviation reported as a subscript. Scores in \textbf{bold} denote the highest result for a given dataset and $\varepsilon$ value.}
    \label{tab:f1}
\end{table*}

\begin{table*}[!ht]
    \centering
\begin{minipage}[t]{0.5\textwidth}\centering
\resizebox{\linewidth}{!}{
\begin{tabular}[t]{ll|cccccccc}
 &  & \multicolumn{8}{c}{$\varepsilon$} \\ \cline{3-10} 
 & \textbf{Yelp} & \multicolumn{1}{c}{0.1} & \multicolumn{1}{c}{0.5} & \multicolumn{1}{c}{1} & \multicolumn{1}{c}{5} & \multicolumn{1}{c}{10} & \multicolumn{1}{c}{15} & \multicolumn{1}{c}{25} & \multicolumn{1}{c}{50} \\ \hline
 \multicolumn{2}{c|}{\textbf{Baseline}} & \multicolumn{8}{c}{Utility: $81.76_{0.8}$ / Adversary: 90.60} \\ \hline
\multicolumn{1}{l|}{} & Utility F1 & $48.1_{0.0}$ & $48.1_{0.0}$ & $48.1_{0.0}$ & $48.1_{0.0}$ & $48.1_{0.0}$ & $48.1_{0.0}$ & $48.1_{0.0}$ & $48.1_{0.0}$ \\
\multicolumn{1}{l|}{S1} & Static F1  & 16.4 & 15.9 & 14.4 & 11.7 & 13.4 & 15.4 & 19.6 & 30.4 \\
\multicolumn{1}{l|}{} & Adaptive F1  & $56.4_{3.6}$ & $58.9_{1.6}$ & $59.7_{3.0}$ & $59.6_{1.2}$ & $59.0_{2.5}$ & $62.1_{2.1}$ & $60.4_{1.3}$ & $59.2_{1.5}$ \\ 
\multicolumn{1}{l|}{} & Relative Gain  & -0.03 & -0.06 & -0.07 & -0.07 & -0.06 & -0.10 & -0.08 & -0.07 \\ \hline
\multicolumn{1}{l|}{} & Utility F1 & $48.1_{0.0}$ & $48.1_{0.0}$ & $48.1_{0.0}$ & $48.1_{0.0}$ & $48.1_{0.0}$ & $50.3_{3.2}$ & $76.5_{4.2}$ & $79.4_{0.3}$ \\
\multicolumn{1}{l|}{S2} & Static F1  & 8.7 & 9.4 & 9.7 & 19.8 & 32.8 & 42.3 & 55.8 & 63.3 \\
\multicolumn{1}{l|}{} & Adaptive F1  & $44.1_{3.4}$ & $44.0_{4.4}$ & $42.9_{2.0}$ & $50.6_{2.3}$ & $55.0_{1.8}$ & $63.6_{0.6}$ & $71.6_{3.2}$ & $82.2_{2.7}$ \\
\multicolumn{1}{l|}{} & Relative Gain  & 0.10 & \textbf{0.10} & 0.11 & \textbf{0.03} & -0.02 & -0.09 & \textbf{0.15} & \textbf{0.06} \\ \hline
\multicolumn{1}{l|}{} & Utility F1  & $48.1_{0.0}$ & $48.1_{0.0}$ & $48.1_{0.0}$ & $48.1_{0.0}$ & $55.2_{10.0}$ & $58.8_{15.2}$ & $69.1_{14.9}$ & $79.4_{1.1}$ \\
\multicolumn{1}{l|}{S3} & Static F1  & 8.9 & 9.4 & 11.0 & 24.8 & 40.9 & 52.2 & 61.2 & 64.3  \\
\multicolumn{1}{l|}{} & Adaptive F1  & $40.9_{5.4}$ & $45.5_{4.1}$ & $39.2_{3.3}$ & $54.9_{0.8}$ & $60.9_{3.8}$ & $67.4_{2.6}$ & $77.5_{3.2}$ & $82.8_{0.8}$ \\
\multicolumn{1}{l|}{} & Relative Gain  & \textbf{0.14} & 0.09 & \textbf{0.16} & -0.02 & \textbf{0.00} & \textbf{-0.02} & -0.01 & \textbf{0.06} \\ \hline
\multicolumn{1}{l|}{} & Utility F1  & $48.1_{0.0}$ & $48.1_{0.0}$ & $48.1_{0.0}$ & $48.1_{0.0}$ & $48.1_{0.0}$ & $48.1_{0.0}$ & $48.1_{0.0}$ & $53.1_{3.7}$ \\
\multicolumn{1}{l|}{S4} & Static F1  & 9.3 & 9.6 & 10.6 & 17.2 & 21.3 & 24.4 & 31.2 & 40.5 \\
\multicolumn{1}{l|}{} & Adaptive F1  & $42.5_{3.7}$ & $45.0_{2.1}$ & $42.0_{7.5}$ & $52.6_{0.5}$ & $56.8_{1.6}$ & $57.4_{2.4}$ & $61.7_{2.2}$ & $66.9_{0.2}$ \\
\multicolumn{1}{l|}{} & Relative Gain  & 0.12 & 0.09 & 0.12 & 0.01 & -0.04 & -0.05 & -0.09 & -0.09
\end{tabular}
}
\end{minipage}\hfill
\begin{minipage}[t]{0.5\textwidth}\centering
\resizebox{0.99\linewidth}{!}{
\begin{tabular}[t]{ll|cccccccc}
 &  & \multicolumn{8}{c}{$\varepsilon$} \\ \cline{3-10} 
 & \textbf{Trustpilot} & \multicolumn{1}{c}{0.1} & \multicolumn{1}{c}{0.5} & \multicolumn{1}{c}{1} & \multicolumn{1}{c}{5} & \multicolumn{1}{c}{10} & \multicolumn{1}{c}{15} & \multicolumn{1}{c}{25} & \multicolumn{1}{c}{50} \\ \hline
 \multicolumn{2}{c|}{\textbf{Baseline}} & \multicolumn{8}{c}{Utility: $98.49_{0.6}$ / Adversary: 68.70} \\ \hline
\multicolumn{1}{l|}{} & Utility F1   & $48.1_{0.0}$ & $50.9_{3.9}$ & $48.1_{0.0}$ & $48.5_{0.6}$ & $48.1_{0.0}$ & $48.1_{0.0}$ & $50.8_{3.8}$ & $68.6_{4.3}$ \\
\multicolumn{1}{l|}{S1} & Static F1  & 58.2 & 58.1 & 57.9 & 57.8 & 57.7 & 58.1 & 58.0 & 60.5 \\
\multicolumn{1}{l|}{} & Adaptive F1  & $58.1_{0.2}$ & $57.9_{0.0}$ & $57.9_{0.9}$ & $58.7_{0.7}$ & $57.6_{1.1}$ & $57.6_{0.8}$ & $57.1_{0.7}$ & $60.5_{2.2}$ \\
\multicolumn{1}{l|}{} & Relative Gain  & -0.36 & \textbf{-0.32} & \textbf{-0.35} & -0.36 & -0.35 & -0.35 & -0.32 & -0.18 \\ \hline
\multicolumn{1}{l|}{} & Utility F1 & $48.1_{0.0}$ & $48.1_{0.0}$ & $48.1_{0.0}$ & $63.9_{12.5}$ & $87.8_{0.7}$ & $94.1_{0.7}$ & $96.7_{0.1}$ & $97.6_{0.2}$ \\
\multicolumn{1}{l|}{S2} & Static F1  & 57.8 & 57.7 & 58.1 & 59.2 & 62.4 & 64.7 & 67.5 & 67.9  \\
\multicolumn{1}{l|}{} & Adaptive F1  & $57.9_{0.0}$ & $57.6_{0.5}$ & $58.5_{0.8}$ & $57.9_{0.0}$ & $64.1_{1.0}$ & $64.3_{4.5}$ & $68.2_{0.5}$ & $68.7_{0.7}$ \\
\multicolumn{1}{l|}{} & Relative Gain  & \textbf{-0.35} & -0.35 & -0.36 & -0.19 & -0.04 & \textbf{0.02} & -0.01 & -0.01\\ \hline
\multicolumn{1}{l|}{} & Utility F1  & $48.1_{0.0}$ & $48.1_{0.0}$ & $48.1_{0.0}$ & $83.4_{1.3}$ & $93.1_{0.6}$ & $94.9_{1.6}$ & $97.8_{0.3}$ & $98.4_{0.2}$ \\
\multicolumn{1}{l|}{S3} & Static F1  & 58.0 & 58.0 & 58.7 & 60.2 & 64.1 & 64.9 & 67.7 & 67.6  \\
\multicolumn{1}{l|}{} & Adaptive F1  & $57.9_{0.0}$ & $57.9_{0.0}$ & $57.9_{0.0}$ & $61.3_{2.0}$ & $66.6_{0.5}$ & $66.6_{0.2}$ & $68.2_{0.1}$ & $65.1_{5.2}$ \\
\multicolumn{1}{l|}{} & Relative Gain  & \textbf{-0.35} & -0.35 & \textbf{-0.35} & \textbf{-0.05} & \textbf{-0.02} & -0.01 & 0.00 & \textbf{0.05} \\ \hline
\multicolumn{1}{l|}{} & Utility F1  & $48.1_{0.0}$ & $48.1_{0.0}$ & $48.1_{0.0}$ & $54.3_{5.1}$ & $78.4_{2.9}$ & $86.3_{0.3}$ & $95.4_{0.7}$ & $95.1_{0.4}$ \\
\multicolumn{1}{l|}{S4} & Static F1  & 57.9 & 58.0 & 58.5 & 59.9 & 62.8 & 64.3 & 66.8 & 67.9 \\
\multicolumn{1}{l|}{} & Adaptive F1  & $58.4_{0.7}$ & $58.7_{0.6}$ & $58.5_{0.7}$ & $59.4_{1.0}$ & $62.9_{1.3}$ & $62.1_{1.5}$ & $62.8_{3.5}$ & $66.9_{2.0}$ \\
\multicolumn{1}{l|}{} & Relative Gain  & -0.36 & -0.37 & -0.36 & -0.31 & -0.12 & -0.02 & \textbf{0.05} & -0.01 \end{tabular}
}
\end{minipage}
\caption{Empirical Privacy Results. The highest \textit{relative gains} (using \textit{adaptive} F1) per $\varepsilon$ are \textbf{bolded}.}
\label{tab:ep}
\vspace{-1em}
\end{table*}

\section{Discussion}
\paragraph{Utility Analysis}
An analysis of the results begins with the strong utility performance of collocation-based perturbation strategies across all tested datasets and $\varepsilon$ values. This effect is especially prominent in the \textsc{SST2} and \textsc{MRPC} tasks. Interestingly, the \textsc{RTE} task presents a challenge for all tested strategies, implying that entailment tasks are more difficult with privatized texts. Nevertheless, the utility loss is dampened with collocation-based methods, particularly at $\varepsilon \ge 1$ (Figure \ref{fig:drop}).

Another intriguing finding comes with the \textsc{CoLA} results, where all strategies struggle to enable any sort of \say{true 
learning} until the $\varepsilon = 10$ threshold. Upon reflection, this particular task may represent the toughest of utility tasks, as the ability to determine the \textit{acceptability} of a given text becomes extremely challenging post-perturbation. Nevertheless, as opposed to S1 (word-level) perturbation, which can never break the worst-case (majority voting) performance, both S2 and S3 are successful in doing this for higher $\varepsilon$ values. One can attribute this to the fact that collocation-based perturbation will still preserve traces of semantic coherence, which is crucial for the \textsc{CoLA} task.

Surprisingly, MST performs poorly in terms of utility as compared to GST. While the exact reason for this would require an in-depth study, we posit that two takeaways can be learned: (1) maximizing PMI might not necessarily be ideal in any case and especially for privatization, and (2) the use of PMI itself may introduce issues, due to the limitations of a frequency-based association measure.

\paragraph{Budget Distribution}
An important discussion arises out of the comparative performance demonstrated by S2 and S3/4. Despite being granted on average a (much) stricter privacy budget, S2 perturbations manage to show strong performance across all tasks, having the highest score in 5 experiment scenarios and otherwise competitive scores. In essence, texts perturbed via S2 hold tighter document-level privacy guarantees than S3/4, yet they are still able to preserve utility better on average than the pure word-level perturbations of S1.

Based on these findings, we hold that further work should be afforded to investigate best practices with budget allocation, including that beyond simple \say{uniform} allocation given a document budget. This becomes more interesting (and potentially complex) with collocations rather than words.

\paragraph{Beyond F1: Similarity and Perplexity}
The \textit{CS} and \textit{AP} metrics also tell an interesting story. On average, collocation-based perturbations always result in privatized texts with higher semantic similarity, even at lower $\varepsilon$ values. The strength of collocations is particularly made clear at higher $\varepsilon$ values, where the gap is quite large. In contrast, the perplexity metric is split based on $\varepsilon$ value: at lower values, word-level perturbations (S1) achieve better scores, whereas at higher scores, S3 prevails. This disparity is insightful, prompting the further study of metric-based evaluations in privacy-preserving NLP. Qualitatively, one can argue that collocation-based perturbations produce much more coherent and readable texts, as showcased in Appendix \ref{sec:appendix}.

\paragraph{The Effect on Privacy}
Analyzing the empirical privacy results also brings insights. As opposed to the disparity in perplexity measurement, a \textit{reverse} trend can be observed with empirical privacy. At lower $\varepsilon$ values, collocation-based perturbations achieve comparable or better privatization against adversaries, whereas this advantage begins to favor word-level approaches at higher privacy budgets. However, the strength of word-level approaches at higher budgets comes with the cost of severely limited utility, as shown by both tasks. 

The \textit{relative gain} results show that in none of the tested scenarios, a positive gain can be observed using word-level perturbations. This comes in contrast to strategies S2-4, which often show positive gains, and achieve the highest relative gain in all but one scenario. These results are promising in the way that MLDP mechanisms can be made practically feasible when leveraging collocations.

As a final analysis, we observe that collocation embedding models enable greater diversity in privatization outputs. Taking the vocabulary of \textsc{deberta-v3-base} (128k tokens), we discover that while only 68,544 unigram tokens from our GST model exist in the vocabulary, \textit{1,248,304} tokens from the model match the vocabulary, i.e., where \textit{every} word exists in the vocabulary. This allows for a wider search space, thus presumably reducing cases where a token is perturbed to itself.

\paragraph{Replication on Other Mechanisms}
We replicate the \textsc{SST2} utility experiments on two other MLDP mechanisms, the Mahalanobis Mechanism \cite{xu2020differentially} and the Vickrey Mechanism \cite{xu2021utilitarian}. These results are shown in Tables \ref{tab:maha} and \ref{tab:vick}. The results mirror those described in this work, albeit with an interesting anomaly observed with the Vickrey Mechanism at lower $\varepsilon$ values. 
We perform this extra analysis as a first step towards generalizing our results to all MLDP mechanisms, in order to investigate the advantages of multi-word rather than single word DP perturbations.

\begin{table}[htbp]
    \centering
    \resizebox{0.9\linewidth}{!}{
\begin{tabular}{c|llll}
Baseline & \multicolumn{4}{c}{$94.33_{0.2}$} \\ \hline
$\varepsilon$ & \multicolumn{1}{c}{0.1} & \multicolumn{1}{c}{1} & \multicolumn{1}{c}{10} & \multicolumn{1}{c}{25} \\ \hline
S1 & $\mathbf{56.0_{3.6}}$ & $\mathbf{56.4_{3.9}}$ & $58.7_{0.7}$ & $64.6_{0.4}$  \\
S2 & $51.1_{0.3}$ & $55.4_{1.7}$ & $76.2_{0.8}$ & $89.5_{0.4}$  \\
S3 & $50.9_{0.1}$ & $54.4_{2.0}$ & $\mathbf{82.6_{0.8}}$ & $\mathbf{91.5_{0.3}}$  \\
S4 & $52.6_{2.4}$ & $53.9_{2.2}$ & $65.6_{0.2}$ & $71.9_{0.7}$
\end{tabular}
}
\caption{\textsc{SST2} Utility Results, using the Mahalanobis Mechanism \cite{xu2020differentially}, with $\lambda = 0.2$.}
\label{tab:maha}
\vspace{-10pt}
\end{table}

\begin{table}[htbp]
    \centering
    \resizebox{0.9\linewidth}{!}{
\begin{tabular}{c|llll}
Baseline & \multicolumn{4}{c}{$94.33_{0.2}$} \\ \hline
$\varepsilon$ & \multicolumn{1}{c}{0.1} & \multicolumn{1}{c}{1} & \multicolumn{1}{c}{10} & \multicolumn{1}{c}{25} \\ \hline
S1 & $\mathbf{83.0_{1.1}}$ & $\mathbf{81.3_{0.1}}$ & $67.4_{0.8}$ & $61.5_{7.1}$   \\
S2 & $50.9_{0.0}$ & $56.1_{1.8}$ & $71.8_{0.4}$ & $78.7_{0.2}$  \\
S3 & $51.0_{0.1}$ & $53.2_{3.2}$ & $\mathbf{74.8_{1.5}}$ & $\mathbf{79.8_{0.6}}$  \\
S4 & $53.0_{1.3}$ & $55.2_{1.6}$ & $64.7_{0.6}$ & $67.3_{1.4}$
\end{tabular}
}
\caption{\textsc{SST2} Utility Results, using the Vickrey Mechanism \cite{xu2021utilitarian}, using the two neighbor variant.}
\label{tab:vick}
\vspace{-10pt}
\end{table}

\section{Conclusion}
In this work, we present an alternative to word-level Metric Differential Privacy, which differs in the way that we tokenize and privatize sensitive input texts on the \textit{collocation} level. We provide two collocation extraction algorithms and their corresponding trained embedding models, showing how word-level MLDP mechanisms can be simply augmented to operate on this higher syntactic level. In our evaluation, we demonstrate the merits of such augmentation, achieving a balance between improved utility, higher semantic coherence, and comparable privacy preservation. 

The results provide researchers with two overarching insights. Using collocations, given the same \textit{overall} budget for a document, we can achieve higher utility while still preserving privacy. At the same time, given the same \textit{per-token} budget, perturbing collocations often outperforms word-by-word privatization. Thus, we make the case that further studies in the field of DP NLP should consider investigating linguistic units outside of the standard word- or sentence-/document-level.

The main limitations of our study come with our reliance on one particular measure for collocation extraction, namely PMI. In addition, we focus on validating our method for the MADLIB mechanism, but do not perform extensive testing on more recent methods. Finally, we base our results on the selected datasets for utility and privacy, whereas this would be well-served to be more extensively tested. As such, we propose the following paths for future work: (1) a focus on collocations and their reliable extraction for DP applications, (2) further work on validating the merits of privatization between the word and sentence level, and (3) deeper investigations into the rigorous evaluation of DP text privatization, with an emphasis on metrics.

\newpage

\section*{Acknowledgments}
The authors would like to thank the anonymous reviewers for their time and feedback and Alexandra Klymenko for her valuable contributions.

\section*{Limitations}
The main limitations regarding our experimental setup include the use of only one metric for automatic collocation extraction. In addition, we do not clean or filter the outputs of the collocation extraction process, outside of our set threshold of $PMI \ge 2$. The effect of performing extra cleaning steps, or by using entirely different collocation extraction methods, remains a point for future work.

Another point is the limited testing in terms of MLDP mechanisms. We decided to test extensively on one mechanism (MADLIB) rather than conduct more limited tests on a variety of mechanisms. Although we provide initial insights into the effect on other mechanisms, further testing is needed.

Finally, we acknowledge the distinction between measured results of \textit{empirical privacy} versus true privacy preservation, and although the former is a good proxy for the latter, there is still work to be done regarding the nature of privacy in textual data. 

\bibliography{custom}

\appendix

\section{Appendix}
\label{sec:appendix}

\begin{table*}[ht!]
    \centering
    \resizebox{\linewidth}{!}{
\begin{tabular}{p{0.07\textwidth}|c|c|c|c|c|c}
 & \multicolumn{6}{c}{\textbf{Tokens}} \\ \cline{2-7} 
 & machinerytrader & mahatma & elise & festival\_itself & wordwide\_market & certificates\_of\_completion \\ \hline
 & crusher\_aggregate\_equipment & gandhiji & anna & whole\_festival & global\_market & course\_certificate \\
\textbf{Most} & portable\_cone\_crusher & swami\_vivekananda & aimee & this\_festival & worldwide\_markets & training\_certificates \\
\textbf{similar} & aggregate\_equipment & bapuji & julia & festival\_weekend & growing\_market & training\_certificate \\
\textbf{tokens} & equipmentmine & babasaheb & sarah & festival\_week & this\_market\_segment & graduation\_certificate  \\
 & bucket\_crusher & savarkar & megan & festival\_period & massive\_market & their\_certificate 
\end{tabular}
}
\caption{Token examples from the \textbf{GST} collocation embedding model. Shown are randomly selected tokens from the model, along with their five most similar tokens in the embedding space.}
\label{tab:collocations}
\end{table*}

\begin{table*}[ht!]
    \centering
    \resizebox{0.85\linewidth}{!}{
\begin{tabular}{lc|llllllll}
 & \multicolumn{1}{l|}{} & \multicolumn{8}{c}{Document Budget ($\varepsilon$)} \\ \hline
\multicolumn{1}{l|}{Dataset} & Avg. Words/Text & \multicolumn{1}{c}{0.1} & \multicolumn{1}{c}{0.5} & \multicolumn{1}{c}{1} & \multicolumn{1}{c}{5} & \multicolumn{1}{c}{10} & \multicolumn{1}{c}{15} & \multicolumn{1}{c}{25} & \multicolumn{1}{c}{50} \\ \hline
\multicolumn{1}{l|}{CoLA} & 7.80 & 0.78 & 3.9 & 7.8 & 38.99 & 77.99 & 116.98 & 194.96 & 389.93 \\
\multicolumn{1}{l|}{MRPC} & 19.54 & 1.95 & 9.77 & 19.54 & 97.72 & 195.44 & 195.44 & 488.6 & 977.21 \\
\multicolumn{1}{l|}{RTE} & 44.48 & 4.45 & 22.24 & 44.48 & 222.41 & 444.82 & 667.23 & 1112.06 & 2224.12 \\
\multicolumn{1}{l|}{SST2} & 8.82 & 0.88 & 4.41 & 8.82 & 44.11 & 88.22 & 132.33 & 220.56 & 441.12 \\
\multicolumn{1}{l|}{Trustpilot} & 52.16 & 5.22 & 26.08 & 52.16 & 260.81 & 521.61 & 782.42 & 1304.03 & 2608.05 \\
\multicolumn{1}{l|}{Yelp} & 186.87 & 18.69 & 93.43 & 186.87 & 934.34 & 1868.68 & 2803.02 & 4671.7 & 9343.41 
\end{tabular}
}
\caption{Document-level budgets. Given our base $\varepsilon$ values, we scale the allocated overall budget per document based on the average token length of documents in each dataset. The resulting budgets are thus shown in the table.}
\label{tab:budgets}
\end{table*}

\paragraph{Collocation Examples}
Table \ref{tab:collocations} presents a sample of six randomly selected tokens from our GST-extracted collocation embedding model, as well as the five nearest neighbors in the space. Note that for any given token, a nearest neighbor need not be the same \say{length} token, i.e., a unigram's nearest neighbor may include bigrams or trigrams.

\paragraph{Document-level Budgets}
As described in Section \ref{sec:util}, to utilize our selected \say{base} $\varepsilon$ values, we scale the privacy budget allotted to each tested dataset. 
In Table \ref{tab:budgets}, we tabulate all document budgets, which are calculated by multiplying the average words per text by the base $\varepsilon$ values.

\paragraph{Examples}
Table \ref{tab:examples_mrpc} shows selected privatization outputs from two datasets using MADLIB with the privatization strategies S1-4. For readability, we strip sentence punctuation marks, and we select five $\varepsilon$ values for illustration. Some inappropriate words have been redacted. 

\begin{table*}[b]
    \footnotesize
    \centering
    \resizebox{0.99\linewidth}{!}{
\begin{tabular}{ccp{0.9\textwidth}}
& \multicolumn{1}{r|}{} & \multicolumn{1}{l}{\textit{Original text:}} \\
\multicolumn{1}{l}{} & \multicolumn{1}{c|}{$\varepsilon$} & this deal makes sense for both companies halla said in a prepared statement \\ \hline
\multicolumn{1}{c|}{} & \multicolumn{1}{c|}{0.1} & ridership rhp [REDACTED] hypothalamus [REDACTED] chiller rm ridership warhead ridership a cyberattacks [REDACTED] \\
\multicolumn{1}{c|}{} & \multicolumn{1}{c|}{0.5} &  chiller chiller ridership lf xp chiller comeuppance [REDACTED] affections rm a [REDACTED] [REDACTED] \\
\multicolumn{1}{c|}{S1} & \multicolumn{1}{c|}{1} &  quercetin chiller cyberattacks unsecure dropkick affections backrest [REDACTED] galaxies transcriptional a comeuppance creole \\
\multicolumn{1}{c|}{} & \multicolumn{1}{c|}{5} & ridership counselor flicker shekels fences sconces rm lidocaine aerodynamics housemates a questionnaires libretto \\
\multicolumn{1}{c|}{} & \multicolumn{1}{c|}{10} & savings hovers occasions dough photographing housemate restrictions renminbi lotion condemning a batsman genocide \\ \hline
\multicolumn{1}{c|}{} & \multicolumn{1}{c|}{0.1} & rbis are worthy especially true who didn animal ’ knockon effect damages that up to 15 alzheimer ’ particularly the case baha ’ s most recent \\
\multicolumn{1}{c|}{} & \multicolumn{1}{c|}{0.5} & enjoyed every dry cleaned domino effect all u multimeter enjoyed every vicious circle vicious cycle audiences who chose marijuana use especially true \\
\multicolumn{1}{c|}{S2} & \multicolumn{1}{c|}{1} & up to 15 especially true potter ’ s publics enjoy reading book consumers ’ found your blog chain of events attempt missed i enjoyed reading forward to reading posted june \\
\multicolumn{1}{c|}{} & \multicolumn{1}{c|}{5} & extract of sample deal that was makes sense poker action both companies 154 receiving means holm shapleigh found across 09  \\
\multicolumn{1}{c|}{} & \multicolumn{1}{c|}{10} & said loudly amazon which makes sense such as gym both firms le film halla said in a prepared statement \\ \hline
\multicolumn{1}{c|}{} & \multicolumn{1}{c|}{0.1} & true even something i could yearold has glad it particularly evident line dry later went particularly the case extremely satisfied publics machine wash change has \\
\multicolumn{1}{c|}{} & \multicolumn{1}{c|}{0.5} & captcha is if nothing true even machinewashable chilling effect nonconference static display is gluten they sleep loved every mile trail gentle cycle \\
\multicolumn{1}{c|}{S3} & \multicolumn{1}{c|}{1} & judged that deet belong on this mitzvot publics weather ’ s blood group its traditions you woke even take especially useful california who \\
\multicolumn{1}{c|}{} & \multicolumn{1}{c|}{5} &  said anna this new agreement makes sense custom construction both sectors marzi 5 responses emily rose announced “ within the garden a prepared statement \\
\multicolumn{1}{c|}{} & \multicolumn{1}{c|}{10} &  any deal that makes sense for both entities thats the truth halla said in a prepared statement \\ \hline
\multicolumn{1}{c|}{} & \multicolumn{1}{c|}{0.1} & t going t think breakfast t see click when t hesitate when i ’ ve look forward is made \\
\multicolumn{1}{c|}{} & \multicolumn{1}{c|}{0.5} & he had ’ d may not will not his wife t be would have t want as t get populations it \\
\multicolumn{1}{c|}{S4} & \multicolumn{1}{c|}{1} &  filed under diameter exchange relationship between tax smaller ° c campaign master very difficult have not like \\
\multicolumn{1}{c|}{} & \multicolumn{1}{c|}{5} & its seems like for plan that seasoned instead said in an easy third floor \\
\multicolumn{1}{c|}{} & \multicolumn{1}{c|}{10} & £ 1 makes sense job search staffers clarinet brokerage firms other said in an excellent immigration and customs \\
\end{tabular}
    }
\caption{Privatization samples from \textsc{MRPC}.} 
\label{tab:examples_mrpc}
\end{table*}

\end{document}